\documentclass[11pt]{article}

\usepackage[preprint]{acl}

\usepackage{times}
\usepackage{latexsym}

\usepackage[T1]{fontenc}

\usepackage[utf8]{inputenc}

\usepackage{microtype}

\usepackage{inconsolata}

\usepackage{graphicx}

\usepackage{amssymb}
\usepackage{microtype}
\usepackage{hyperref}
\usepackage{url}
\usepackage{booktabs}
\usepackage{amsmath}
\usepackage{enumitem}
\usepackage{graphicx}
\usepackage{lineno}
\usepackage{xspace}
\usepackage{algorithm}
\usepackage{algpseudocode}
\usepackage{array}
\usepackage{booktabs}
\usepackage{longtable}
\usepackage{arydshln}
\usepackage{wrapfig}
\usepackage{caption}
\usepackage{bm}
\usepackage{subcaption} 
\usepackage{tcolorbox}

\usepackage{multirow}

\definecolor{darkblue}{rgb}{0, 0, 0.5}
\hypersetup{colorlinks=true, citecolor=darkblue, linkcolor=darkblue, urlcolor=darkblue}
\definecolor{viridis3}{RGB}{33, 145, 140}

\newcommand{\call}[1]{\textcolor{cyan}{\texttt{<call>}} #1 \textcolor{cyan}{\texttt{</call>}}}
\newcommand{\info}[1]{\textcolor{brown}{\texttt{<resp>}} #1 \textcolor{brown}{\texttt{</resp>}}}
\newcommand{\answer}[1]{\textcolor{purple}{\texttt{<ans>}} #1 \textcolor{purple}{\texttt{</ans>}}}

\newcommand{\Ours}{\textsc{CoRL}\xspace}

\title{Controlling Performance and Budget of a Centralized Multi-agent LLM System with Reinforcement Learning}

\author{Bowen Jin$^{12}$, TJ Collins$^2$, Donghan Yu$^2$, Mert Cemri$^{23}$, Shenao Zhang$^{24}$, Mengyu Li$^2$, \\ \textbf{Jay Tang$^2$, Tian Qin$^{25}$, Zhiyang Xu$^{26}$,  Jiarui Lu$^2$, Guoli Yin$^2$, Jiawei Han$^1$, Zirui Wang$^2$} \\
  $^1$University of Illinois at Urbana-Champaign, $^2$Apple, $^3$University of California, Berkeley \\
  $^4$Northwestern University, $^5$Harvard University, $^6$Virginia Tech \\
  \texttt{\small \{bowenj4,hanj\}@illinois.edu, cemri@berkeley.edu, shenao@u.northwestern.edu} \\
  \texttt{\small \{tjcollins,donghan\_yu,mengyu\_li2,x\_tang,tian\_qin2,zhiyang\_xu2,jiarui\_lu,gyin,zirui\_wang3\}@apple.com} \\
  }

\begin{document}
\maketitle
\begin{abstract}
Large language models (LLMs) exhibit complementary strengths across domains and come with varying inference costs, motivating the design of multi-agent LLM systems where specialized models collaborate efficiently. 
Existing approaches predominantly rely on decentralized frameworks, which invoke multiple LLMs for every input and thus lead to substantial and uncontrolled inference costs.
In this work, we introduce a centralized multi-LLM framework, where a controller LLM selectively coordinates a pool of expert models in a cost-efficient and cost-controllable manner. 
We formulate this coordination problem as reinforcement learning with dual objectives: maximizing task performance while minimizing the overall inference cost.
In addition, we expect the multi-agent system to have adapted behavior with different budget conditions during inference.
To this end, we propose \Ours, a reinforcement learning framework that optimizes the performance–cost trade-off in a controllable multi-budget setting. 
Experiments on four diverse benchmarks demonstrate that \Ours enables a single system to surpass the best expert LLM under high-budget settings, while maintaining strong performance in more economical low-budget modes, highlighting the effectiveness of centralized coordination for scalable and cost-efficient multi-agent LLM systems.
\end{abstract}

\section{Introduction}

Large language models (LLMs) \cite{zhao2023survey,achiam2023gpt,comanici2025gemini} exhibit complementary strengths across domains, offering unique capabilities at different inference costs.
For example, coding LLMs \cite{jiang2024survey,guo2024deepseek} excel in software engineering tasks, reasoning-focused LLMs \cite{guo2025deepseek,jaech2024openai} solve complex mathematical problems, and multimodal LLMs \cite{liu2023visual,li2023blip,chen2025blip3} process diverse data types effectively—often at very different inference costs.
This heterogeneity in both capabilities and costs naturally motivates the design of multi-LLM systems, where specialized models collaborate to tackle complex and diverse tasks more effectively and economically than any single model.

Many existing works on multi-LLM systems \cite{subramaniam2025multiagent,park2025maporl} adopt a decentralized framework, where the input question is simultaneously sent to all LLMs for discussion and debate. While this strategy can improve task accuracy through collaboration, it often leads to substantial inference costs because each model must perform a full rollout for every input. For example, the OpenAI o1 \cite{jaech2024openai} model incurs a cost of \$15.00 per million input tokens and \$60.00 per million output tokens, making repeated calls to multiple large models prohibitively expensive at scale.

To address this, it is crucial to design a \textit{cost-efficient} and \textit{cost-controllable} multi-LLM system. Such a system should (1) intelligently and selectively dispatch queries to the most suitable and economical expert model, rather than defaulting to the strongest (and typically most expensive) model, and (2) enable dynamic control over inference behavior—ranging from low-budget to high-budget modes—while (3) allowing the cheap controller to independently handle queries when possible, thereby avoiding unnecessary external model calls.

In this work, we propose a reinforcement learning–based framework for cost-controllable multi-LLM coordination. We adopt a centralized architecture where specialized LLMs collaborate under the guidance of a controller LLM \cite{fourney2024magentic, tao2024magis}. Given an input question, the controller first decides whether to answer it directly or to decompose it into sub-queries, which are selectively dispatched to expert models as needed. The entire system is trained using RL with two complementary reward signals: a task performance reward and a cost reward, enabling the controller to learn budget-aware decision-making.
To further support adaptive inference under varying budget constraints, we introduce a cost-controllable training strategy, where samples associated with different budget levels are conditioned on distinct system prompts and receive budget-specific rewards. This allows the resulting system to operate flexibly across multiple budget regimes at inference time—achieving strong performance in high-budget modes while remaining economical in low-budget settings.

In summary, our contributions are threefold:
\begin{itemize}[leftmargin=*,nosep]
    \item We introduce and formalize the problem of training a cost-controllable multi-LLM system, which aims to jointly optimize task performance and inference cost under different budget constraints.
    \item We develop a reinforcement learning framework that trains a controller LLM to coordinate multiple expert models in a cost-effective and controllable manner, enabling dynamic budget-aware decision-making during inference.
    \item We conduct extensive experiments on four diverse datasets, demonstrating that our approach achieves strong performance in high-budget modes while maintaining cost efficiency in low-budget settings.
\end{itemize}

\section{Related Works}

\paragraph{Training LLMs with Reinforcement Learning.}

Reinforcement learning (RL) \citep{kaelbling1996reinforcement} is a learning paradigm in which an agent learns to make sequential decisions by interacting with an environment \citep{sutton1999reinforcement}. 
RL was adopted to large language model (LLM) tuning by \citet{ouyang2022training} through reinforcement learning from human feedback (RLHF) \citep{kaufmann2023survey}. 
RLHF trains a reward model from human preference data \citep{lambert2024rewardbench}, which then guides RL-based policy optimization of the LLM, typically via Proximal Policy Optimization (PPO). 
To address the optimization challenge of PPO, several alternative algorithms have been proposed. 
Group Relative Policy Optimization (GRPO) \citep{shao2024deepseekmath} removes the need for a separate critic by estimating baselines from group scores. 
RLOO \citep{ahmadian2024back} simplifies optimization through a REINFORCE-style \citep{williams1992simple} objective without auxiliary models. 
DAPO \citep{yu2025dapo} introduces four techniques to improve RL performance in long chain-of-thought (CoT) scenarios, while GSPO \citep{zheng2025group} stabilizes training by computing importance ratios based on sequence likelihood and applying sequence-level clipping and reward assignment.
Beyond alignment, RL has been increasingly applied to training LLM-based search agents \citep{jin2025search, jin2025empirical}. However, the use of RL to train cost-efficient and cost-controllable multi-LLM systems remains largely unexplored, despite its potential to optimize both performance and inference budget simultaneously.

\paragraph{Multi-LLM systems.}

Multi-LLM systems have shown strong potential by coordinating multiple specialized expert models to tackle complex tasks \citep{li2024survey, sun2024llm, cemri2025multi}. Existing approaches fall into two categories: prompting-based and fine-tuning-based methods. Prompting-based methods, such as AutoGen \citep{wu2024autogen} and Camel \citep{li2023camel}, use carefully designed role-playing prompts to enable communication and collaboration among multiple LLMs without additional training. Fine-tuning-based methods, including multi-LLM fine-tuning \citep{subramaniam2025multiagent}, optimize collaboration through training. For example, they propose self-improvement via multi-LLM debate and majority-vote supervision. MasRouter \citep{yue2025masrouter} further introduces cascaded controller networks for collaboration mode determination, role allocation, and routing.
Reinforcement learning has also been applied to multi-LLM coordination. MLPO \citep{estornell2025train} trains a leader LLM through RL based on responses from other agents, while MAPoRL \citep{park2025maporl} removes the leader and enables multi-LLM debate via RL rollouts. 
However, existing RL methods focus solely on task performance and overlook cost as a primary optimization objective. 
Router-R1 \citep{zhang2025router} is the most related work, but it mainly targets performance improvement and does not consider controllable behavior under different inference budget modes, which is the central focus of this work.

\section{Framework}

\subsection{Overall Framework}\label{sec:rollout}

\begin{figure}[t] 
    \centering
    \includegraphics[width=0.8\linewidth]{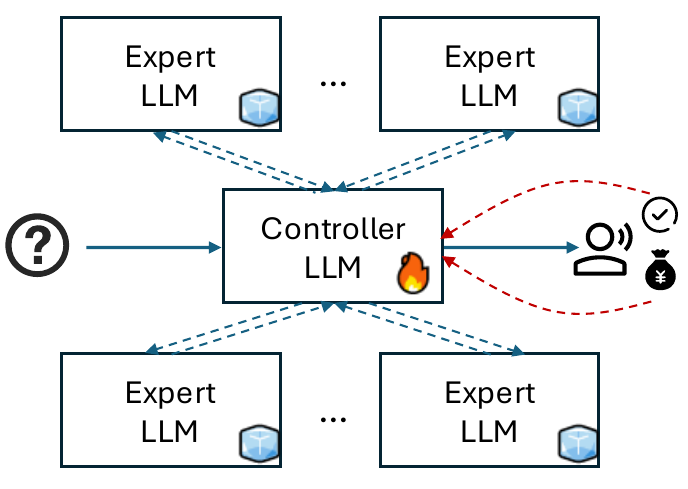} 
    \caption{Overview of \Ours. We adopt a centralized multi-LLM architecture, where a controller LLM coordinates interactions with multiple expert LLMs. The system is trained via RL with dual rewards for task performance and multi-level query cost, while only the controller LLM is optimized for efficiency.}
    \label{fig:main} 
\end{figure}

Following \citet{fourney2024magentic, tao2024magis}, we adopt a centralized multi-LLM framework (as shown in Figure \ref{fig:main}), where there is a controller LLM responsible for analyzing the input problem, decomposing it into subproblems, and feeding them into other expert LLMs, as needed.

To be specific, given the controller LLM $\pi_\theta$ and a pool $\Gamma$ of $K$ other expert LLMs, $\Gamma = \{\pi_{\gamma_i}\}_{1\leq i \leq K}$, the multi-LLM system rollout $\pi_{\theta}(\cdot|\bm{x};\Gamma)$ is done by:
\begin{gather}
    \pi_{\theta}(\cdot|\bm{x};\Gamma) = \prod^M_{j=1} \pi_\theta(\bm{y}^j_\theta|\bm{x},\bm{y}^{j-1}) \pi_{\gamma^{j}}(\bm{y}^j_{\gamma^{j}}|g(\bm{y}^j_\theta)), \\\
    \gamma^{j} = f(\bm{y}^j_\theta)
\end{gather}
where $M$ is the number of interaction rounds between the controller and the expert LLMs. 
$\bm{x}$ is the input prompt and $\bm{y}^j_\theta$ is the $j$-th iteration controller response should include controller's reasoning, expert selection and queries.
$\gamma^{j}$ is selected by a function $f(\bm{y}^j_\theta)$ which maps the controller $\pi_\theta$'s current iteration decision $\bm{y}^j_\theta$ to one expert LLM $\pi_{\gamma^{j}} \in \Gamma$. 
$g(\bm{y}^j_\theta)$ is a function which extracts the queries from $\bm{y}^j_\theta$ that will be sent to $\pi_{\gamma^{j}}$.
Here, both $f(\cdot)$ and $g(\cdot)$ are rule-based string parsing functions.
$\bm{y}^j$ is the rollout trajectory after the $j$-th iteration which is the concatenation of both the previous rounds controller's decisions and expert LLM responses:
\begin{gather}
    \bm{y}^j = \bm{y}^{j-1} + \bm{y}_\theta^j + \bm{y}_{\gamma^{j}}^j
\end{gather}
The detailed calculation of $\pi_{\theta}(\cdot|\bm{x};\Gamma)$ can be found in Algorithm \ref{alg:llm-rollout}.

\begin{algorithm}[t]
\caption{Centralized Multi-LLM Rollout.}
\label{alg:llm-rollout}
\begin{algorithmic}[1]
\Require Input query \( \bm{x} \), a controller LLM \( \pi_{\theta} \), a LLM pool $\Gamma = \{\pi_{\gamma_i}\}_{1\leq i \leq K}$, the maximum action budget \( M \).
\Ensure Final response \( \bm{y} \).

\State Initialize rollout sequence \( \bm{y} \gets \emptyset \)
\State Initialize action count \( j \gets 1 \)

\While{\( j < M \)}
    \State \( \bm{y}_\theta^j \gets \emptyset \) 
    \While{True}
    \State \( y_t \sim \pi_{\theta}(\cdot \mid \bm{x}, \bm{y} + \bm{y}_\theta^j) \)
    \State \( \bm{y}_\theta^j \gets \bm{y}_\theta^j + y_t \)
    \If{\( y_t \) in [\textcolor{cyan}{\texttt{</call>}}, \textcolor{purple}{\texttt{</ans>}}, \texttt{<eos>}]}
        break
    \EndIf
    \EndWhile

    \State \( \bm{y}  \gets  \bm{y} + \bm{y}_\theta^j \)
    \If{\call{} detected in \( \bm{y}_\theta^j \)}
        \State \( \bm{q}^j \gets \text{Parse}_g(\bm{y}_\theta^j) \)
        \State \( \gamma^{j} \gets \text{Parse}_f(\bm{y}_\theta^j) \)
        \State \( \bm{y}^j_{\gamma^{j}} = \pi_{\gamma^{j}}(\bm{q}^j) \)
        \State \( \bm{y}  \gets  \bm{y} + \info{\bm{y}^j_{\pi_{\gamma^{j}}}}  \)
    \ElsIf{\answer{} detected in \( \bm{y}^\theta_j \)}
        \State \textbf{return} final generated response \( \bm{y} \)
    \EndIf

    \State Increment action count \( j \gets j + 1 \)
\EndWhile

\State \textbf{return} final generated response \( \bm{y} \)
\end{algorithmic}
\end{algorithm}

\subsection{Training the Controller LLM with Reinforcement Learning}

We formulate the reinforcement learning objective as follows:
\begin{equation}\label{eq:rl-retriever}
\begin{gathered}
    \max_{\pi_\theta} \mathbb{E}_{\bm{x} \sim \mathcal{D}, \bm{y} \sim \pi_{\theta}(\cdot|\bm{x};\Gamma)} 
\left[ r_{\phi}(\bm{x}, \bm{y}) \right] \notag \\
- \beta \mathbb{D}_{\text{KL}} \left[ \pi_{\theta}(\cdot|\bm{x};\Gamma) \,||\, \pi_{\text{ref}}(\cdot|\bm{x};\Gamma) \right],
\end{gathered}
\end{equation}
where $\pi_{\text{ref}}$ is the reference LLM, $r_{\phi}$ is the reward function and $\mathbb{D}_{\text{KL}}$ is KL-divergence measure.
\( \bm{x} \) denote input samples drawn from the dataset \( \mathcal{D} \), and \( \bm{y} \) represent the generated outputs interleaved with multi-LLM calling responses as discussed in Section \ref{sec:rollout}. 
Unlike prior RL approaches that primarily rely on the policy LLM $\pi_{\theta}(\cdot \mid \bm{x})$ to generate rollout sequences \citep{rafailov2023direct, ouyang2022training}, our framework explicitly incorporates multi-LLM responses via $\pi_{\theta}(\cdot \mid \bm{x}; \Gamma)$.

To simplify the optimization of the multi-LLM system, we only update the parameter of the controller LLM and leave the expert LLMs frozen. We leave the study of training all the LLMs in such a system for future work.

Following \citet{jin2025search}, we adopt PPO \citep{schulman2017proximal} as the reinforcement learning algorithm and apply masking for tokens from the expert LLMs when calculating the learning objective for policy controller update.
To be specific, the objective is shown as follows:
\begin{equation}\label{eq:ppo}
\begin{gathered}
\mathcal{J}_{\mathrm{PPO}}(\theta)
= \mathbb{E}_{x \sim \mathcal{D},\, y \sim \pi_{\text{old}}(\cdot \mid x;\,\Gamma)}
\bigg[
\frac{1}{\sum_{t=1}^{|y|} I(y_t)} \\
\sum_{\substack{t=1\\ I(y_t)=1}}^{|y|}
\min \Big(
r_t(\theta) A_t,\;
\operatorname{clip}\big(r_t(\theta), 1-\epsilon, 1+\epsilon\big) A_t
\Big)
\bigg],
\\
r_t(\theta)
= \frac{\pi_{\theta}(y_t \mid x, y_{<t};\,\Gamma)}
{\pi_{\text{old}}(y_t \mid x, y_{<t};\,\Gamma)},
\end{gathered}
\end{equation}
where $I(y_t)$ is the token loss masking operation such that $I(y_t)=1$ if $y_t$ is a token generated by the controller and $I(y_t)=0$ if $y_t$ is a token generated by other expert LLMs.
The term \( \epsilon \) is a clipping-related hyperparameter introduced in PPO to stabilize training. 
The advantage estimate \( A_t \) is computed using Generalized Advantage Estimation (GAE) \citep{schulman2015high}, based on future rewards \( \{ r_{\geq t} \} \) and a learned value function \( V_{\phi} \).





\begin{figure*}[t]
    \centering
    \begin{subfigure}[b]{0.24\textwidth}
        \centering
        \includegraphics[width=\textwidth]{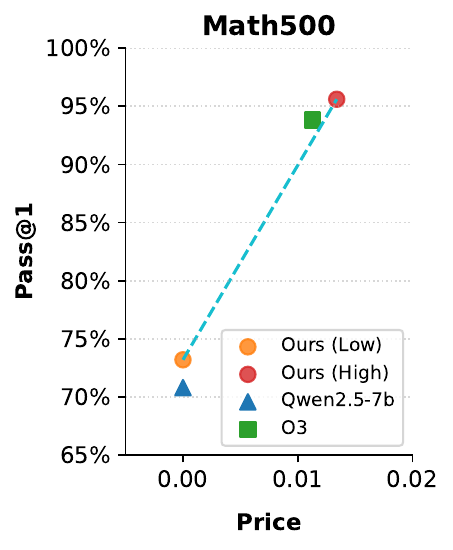}
        \caption{MATH500}
        \label{fig:subfig_b}
    \end{subfigure}
    \hfill
    \begin{subfigure}[b]{0.24\textwidth}
        \centering
        \includegraphics[width=\textwidth]{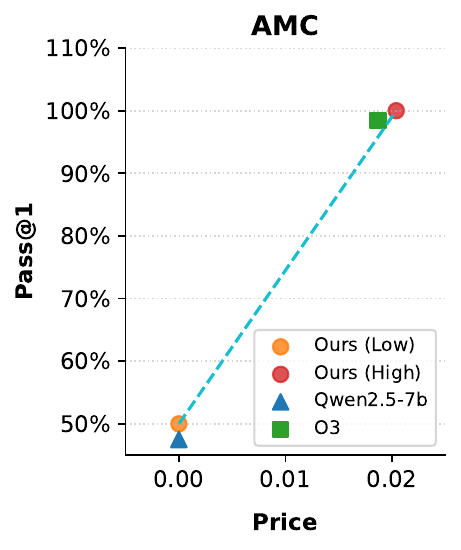}
        \caption{AMC 2023}
        \label{fig:subfig_c}
    \end{subfigure}
    \hfill
    \begin{subfigure}[b]{0.24\textwidth}
        \centering
        \includegraphics[width=\textwidth]{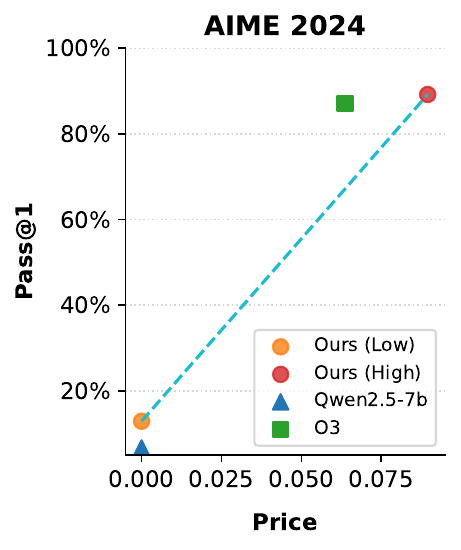}
        \caption{AIME 2024}
        \label{fig:subfig_d}
    \end{subfigure}
    \hfill
    \begin{subfigure}[b]{0.24\textwidth}
        \centering
        \includegraphics[width=\textwidth]{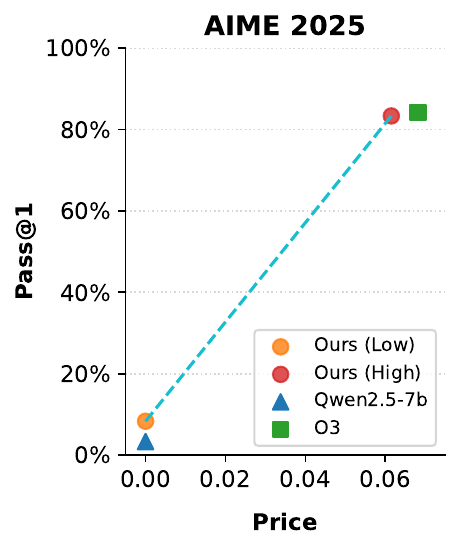}
        \caption{AIME 2025}
        \label{fig:subfig_e}
    \end{subfigure}

    \caption{\textbf{Performance–cost trade-off in the two-LLM system.} x-axis: per-query cost; y-axis: task performance (higher is better). In the \emph{low-budget} mode, \Ours primarily answers with the controller (Qwen2.5-7B-Instruct) and surpasses the controller-alone baseline on all four datasets. In the \emph{high-budget} mode, \Ours leverages the expert (o3) and exceeds the single o3 baseline on three of the four datasets.}
    \label{fig:two-llm-main}
\end{figure*}

\subsection{Controlling the Balance between Performance and Cost}

\paragraph{Reward Design.}

The reward consists of two parts: (1) performance reward $r_p(\bm{x}, \bm{y})$ and (2) cost reward $r_c(\bm{y})$.
For the performance reward $r_p(\bm{x}, \bm{y})$, we follow \cite{fang2025thinkless} and use the accuracy of the prediction as the reward.
For the cost reward $r_c(\bm{y})$, it is calculated based on comparing the cost $c(\bm{y})$ for obtaining the rollout $\bm{y}$ with a predefined cost budget $B$.
\begin{gather}
    r_c(\bm{y}) =
    \begin{cases}
        1, & \text{if } c(\bm{y}) \leq B \\
        0, & \text{if } c(\bm{y}) > B .
    \end{cases}
\end{gather}
The final reward function is a multiplication of the performance reward and cost reward:
\begin{gather}
    r_{\phi}(\bm{x}, \bm{y}) = r_p(\bm{x}, \bm{y}) \cdot r_c(\bm{y}).
\end{gather}
To be specific, the reward is calculated as
\begin{gather}
    r_{\phi}(\bm{x}, \bm{y}) =
    \begin{cases}
        r_p(\bm{x}, \bm{y}), & \text{if } c(\bm{y}) \leq B \\
        0, & \text{if } c(\bm{y}) > B .
    \end{cases}
\end{gather}

We do not include format reward for simplicity, since it does not affect the final performance much for instruction-tuned LLMs, as shown in \cite{jin2025empirical}.

\paragraph{Controlling for different budget levels.}

When deploying a trained multi-LLM system, users may have highly diverse budget constraints.
As a result, it is important that the multi-LLM system learn to be controllable with respect to the budget modes input during training.
To achieve this, we explicitly mention the conditional budget inside the input prompt during training, similar to:
\begin{gather}
    \bm{x} = \text{Concate}(\bm{x}, \text{``Answer under budget \textcolor{red}{a}.''}),
\end{gather}
where $a$ is a budget level that can take on the values \emph{low}, \emph{medium}, or \emph{high}.
During training, we randomly assign different budget mode conditions for different samples and conduct multi-budget mode training.
In addition, training samples with different budget modes will be assigned different cost budgets $B$ for reward calculation, \textit{i.e.}, smaller $B$ for low budget condition and bigger $B$ for high budget condition.

\begin{table}[t]
\centering 
\caption{Training and evaluation datasets.}\label{tb-data-size}
\renewcommand\arraystretch{1.0}
\fontsize{8}{10}\selectfont
\begin{tabular}{@{}lclc@{}}
\toprule
\textbf{Dataset} & \textbf{Domain} & \textbf{Type} & \textbf{\# Samples} \\ 
\midrule
Deepscaler & Math & Training & 40,315 \\
\midrule
MATH500   & Math & Evaluation & 500 \\
AMC 2023  & Competition & Evaluation & 40 \\
AIME 2024 & Competition & Evaluation & 30 \\
AIME 2025 & Competition & Evaluation & 30 \\
\bottomrule
\end{tabular}
\end{table}

\section{Experiments}

\subsection{Datasets \& Evaluation}

We conduct experiments on math reasoning tasks.
We adopt Deepscaler \cite{deepscaler2025} as the training data while reserving MATH500 \cite{lightman2023let}, AMC2023 \cite{amc2023_dataset}, AIME2024 \cite{aime2024_dataset}, and AIME2025 \cite{aime2025_dataset} as the evaluation datasets.
The statistics of the datasets can be found in Table \ref{tb-data-size}.
Given that the size of AMC2023, AIME2024, and AIME2025 is small, we follow \cite{ye2025aimepreview}, sample 8 times per question, and take the average score as the final score. 
For the evaluation of cost, we show the \$ needed per query or for the whole dataset.

\begin{table*}[t]
\centering
\caption{\textbf{Pass@1 and cost across four test sets.} Each pair of bars shows Pass@1 and total cost (in \$) for the entire dataset. (1) In the \emph{high-budget} mode, \Ours outperforms the best individual expert model across all four datasets. (2) \Ours exhibits \textbf{controllable behavior} across budget modes, achieving strong performance–cost trade-offs.}
\label{tab:four-llm}
\renewcommand{\arraystretch}{1.1}
\setlength{\tabcolsep}{4pt}
\begin{tabular}{lcccccccc}
\toprule
\multirow{2}{*}{\textbf{Model}} &
\multicolumn{2}{c}{\textbf{MATH 500}} &
\multicolumn{2}{c}{\textbf{AMC 2023}} &
\multicolumn{2}{c}{\textbf{AIME 2024}} &
\multicolumn{2}{c}{\textbf{AIME 2025}} \\
\cmidrule(lr){2-3} \cmidrule(lr){4-5} \cmidrule(lr){6-7} \cmidrule(lr){8-9}
 & Pass@1 & Cost & Pass@1 & Cost & Pass@1 & Cost & Pass@1 & Cost \\
\midrule
\textbf{Controller} \\
Qwen2.5-7B-it & 0.708 & -- & 0.475 & -- & 0.067 & -- & 0.033 & -- \\
\midrule
\textbf{Expert LLMs} \\
GPT-4.1         & 0.872 & 3.581   & 0.869 & 0.540   & 0.492 & 0.753   & 0.371 & 0.785 \\
GPT-4.1-nano    & 0.816 & 0.192   & 0.678 & 0.031   & 0.233 & 0.041   & 0.188 & 0.037 \\
o3              & 0.938 & 5.642   & 0.984 & 0.747   & 0.871 & 1.911   & 0.842 & 2.045 \\
Random Assign
                & 0.880 & 3.138   & 0.841 & 0.439   & 0.517 & 0.902   & 0.496 & 0.956 \\
\midrule
\textbf{Ours} \\
Low            & 0.900 & 4.650   & 0.938 & 0.660   & 0.733 & 1.506   & 0.725 & 1.983 \\
Medium         & 0.940 & 6.250   & 0.981 & 0.820   & 0.842 & 1.821   & 0.792 & 2.328 \\
High           & \textbf{0.958} & 5.870   & \textbf{0.997} & 0.860   & \textbf{0.877} & 1.913   & \textbf{0.867} & 2.399 \\
\bottomrule
\end{tabular}
\end{table*}

\subsection{Experimental Settings}

We use Qwen2.5-7B-Instruct \citep{team2024qwen2} as the controller LLM and treat three GPT-family models as external experts: o3 (Medium) \citep{openai_o3_system_card_2025_misc}, GPT-4.1, and GPT-4.1-nano \citep{openai_gpt4_1_2025_misc}. 
These experts span a consistent accuracy–cost spectrum on math-reasoning tasks, with both performance and inference price ordered as o3 > GPT-4.1 > GPT-4.1-nano, enabling controlled evaluation across different budget modes.

We set the learning rate of the policy LLM to 1e-6 and that of the value LLM to 1e-5. The maximum training step is step to be 1000, with warm-up ratios of 0.285 and 0.015 for the policy and value models, respectively. We use Generalized Advantage Estimation (GAE) with parameters $\lambda = 1$ and $\gamma = 1$.
More details can be found in Appendix \ref{apx:setting}.

\subsection{Single Expert LLM Results}

We first evaluate our method on a two-LLM system, using Qwen2.5-7B-Instruct \citep{team2024qwen2} as the controller and o3 \citep{openai_o3_system_card_2025_misc} as the expert LLM. We consider two budget levels: low budget, where the controller (Qwen2.5-7B) is encouraged to solve problems independently, and high budget, where the controller is encouraged to consult the expert LLM (o3). The corresponding system prompts for the two budget levels are provided in Appendix~\ref{apx:two-llm-prompt}. 
Training is run for 250 steps until convergence, and more details can be found in Appendix \ref{apx:setting}.

Figure~\ref{fig:two-llm-main} reports the performance and cost of (a) Qwen2.5-7B-Instruct alone, (b) o3 alone, (c) our method under the low-budget mode, and (d) our method under the high-budget mode. The results show that:
(1) With our tailored training design, \Ours exhibits clearly distinct and controllable behaviors under different budget levels;
(2) In the low-budget mode, \Ours primarily relies on the controller to solve problems independently and consistently outperforms Qwen2.5-7B across all four datasets;
(3) In the high-budget mode, \Ours effectively leverages the expert LLM and achieves even higher performance than o3 on three out of four datasets via adapted query phrasing. On AIME 2025, although the performance matches that of o3, \Ours attains a better performance–cost trade-off, as indicated by the dashed line in the figure.

\begin{figure*}[t]
    \centering
    \begin{subfigure}[b]{0.32\textwidth}
        \centering
        \includegraphics[width=\textwidth]{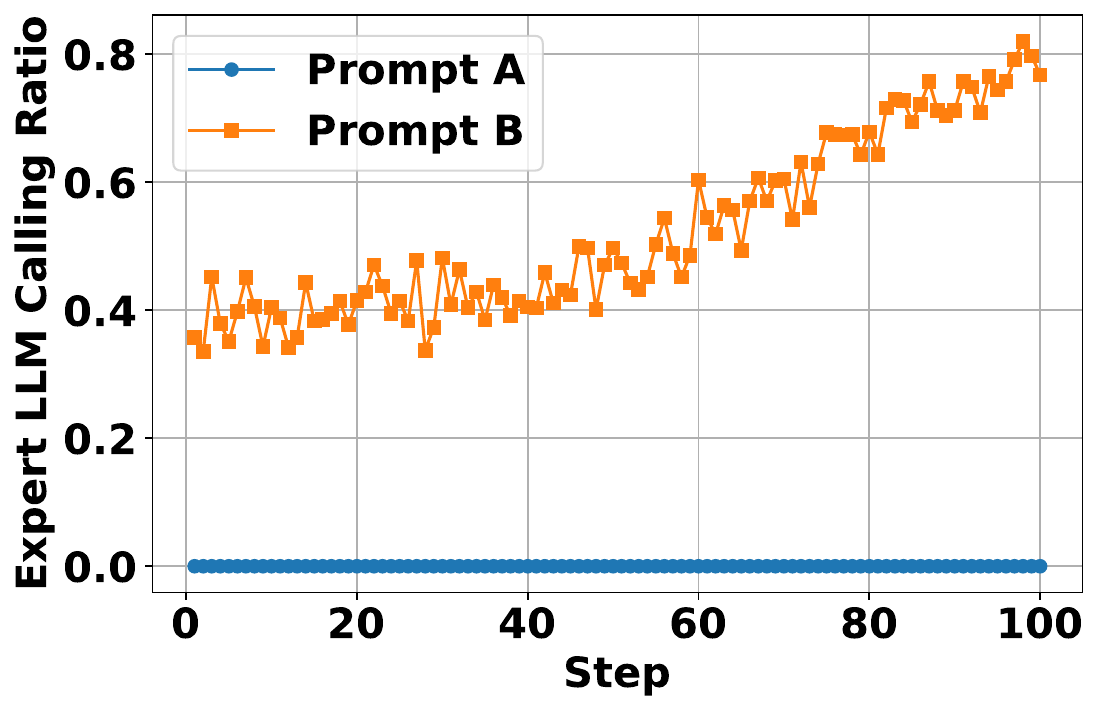}
        \caption{Low Budget Mode}
        \label{fig:llm_ratio_low}
    \end{subfigure}\hfill
    \begin{subfigure}[b]{0.32\textwidth}
        \centering
        \includegraphics[width=\textwidth]{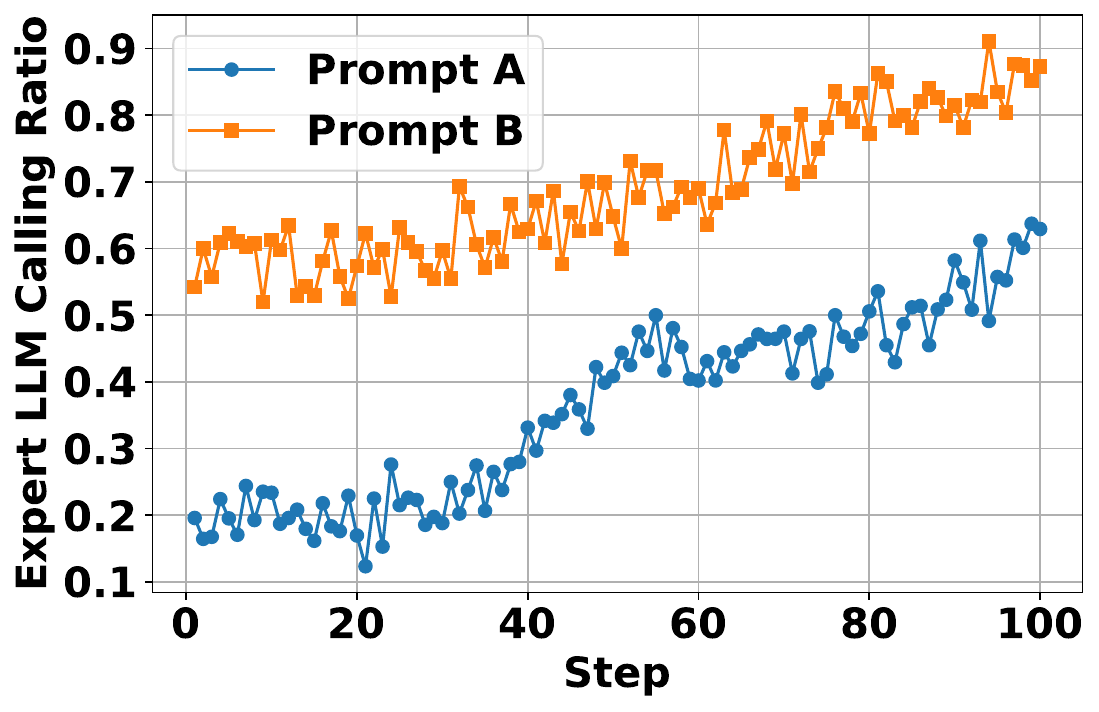}
        \caption{Medium Budget Mode}
        \label{fig:llm_ratio_medium}
    \end{subfigure}\hfill
    \begin{subfigure}[b]{0.32\textwidth}
        \centering
        \includegraphics[width=\textwidth]{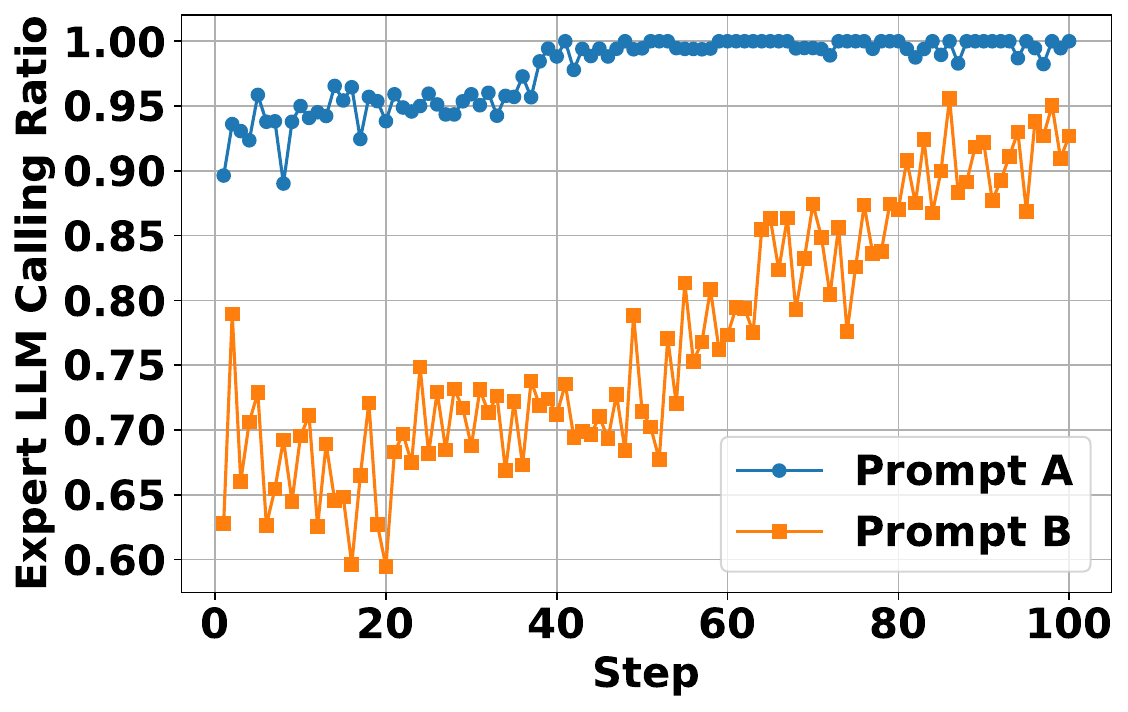}
        \caption{High Budget Mode}
        \label{fig:llm_ratio_high}
    \end{subfigure}
    \caption{\textbf{Ratio of expert LLM calls under different budget modes.} Prompt A and Prompt B correspond to more constrained and more flexible system prompts, respectively. (1) For both prompt types, the expert call ratio follows the expected order of \emph{low} < \emph{medium} < \emph{high}, learned through RL. (2) Overall, the ratio of expert calls increases as training progresses.}
    \label{fig:llm-ratio}
\end{figure*}

\subsection{Multiple Expert LLM Results}

We next evaluate our method on a four-LLM system, where Qwen2.5-7B-Instruct serves as the controller and GPT-4.1, GPT-4.1-nano, and o3 act as expert LLMs. We consider three budget levels: (1) a low-budget mode, where the controller is encouraged to prioritize cheaper experts; (2) a medium-budget mode, where the controller is expected to balance between inexpensive and expensive experts; and (3) a high-budget mode, where the controller is allowed to rely more on the most capable but costly experts. The corresponding system prompts for these modes are provided in Appendix~\ref{apx:four-llm-prompt}. 
Training is run for 200 steps until convergence, and more details can be found in Appendix \ref{apx:setting}.

Table~\ref{tab:four-llm} presents the performance and cost of (a) Qwen2.5-7B-Instruct alone, (b) individual GPT-4.1/GPT-4.1-nano/o3 experts, (c) random routing among the three experts, and (d) \Ours under low-, medium-, and high-budget modes. The results show that:
(1) In the high-budget mode, \Ours outperforms even the best individual expert model across all four datasets;
(2) \Ours substantially surpasses the random routing baseline, indicating that it learns meaningful routing strategies and captures semantic signals that guide expert selection;
(3) \Ours exhibits controllable behavior across budget modes, with the low-budget mode achieving decent performance at minimal cost, and the high-budget mode attaining the best performance at higher cost.

\begin{figure*}[t]
    \centering
    \begin{subfigure}[b]{0.32\textwidth}
        \centering
        \includegraphics[width=\textwidth]{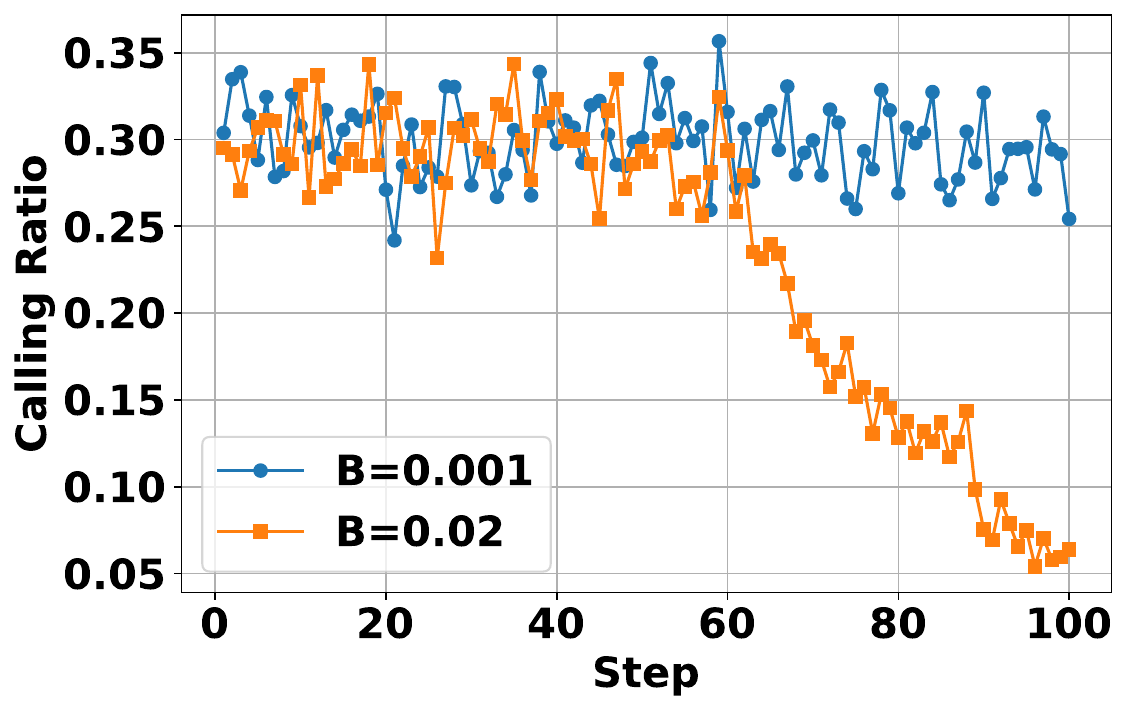}
        \caption{GPT-4.1-nano}
        \label{fig:llm_ratio_gpt41nano}
    \end{subfigure}\hfill
    \begin{subfigure}[b]{0.32\textwidth}
        \centering
        \includegraphics[width=\textwidth]{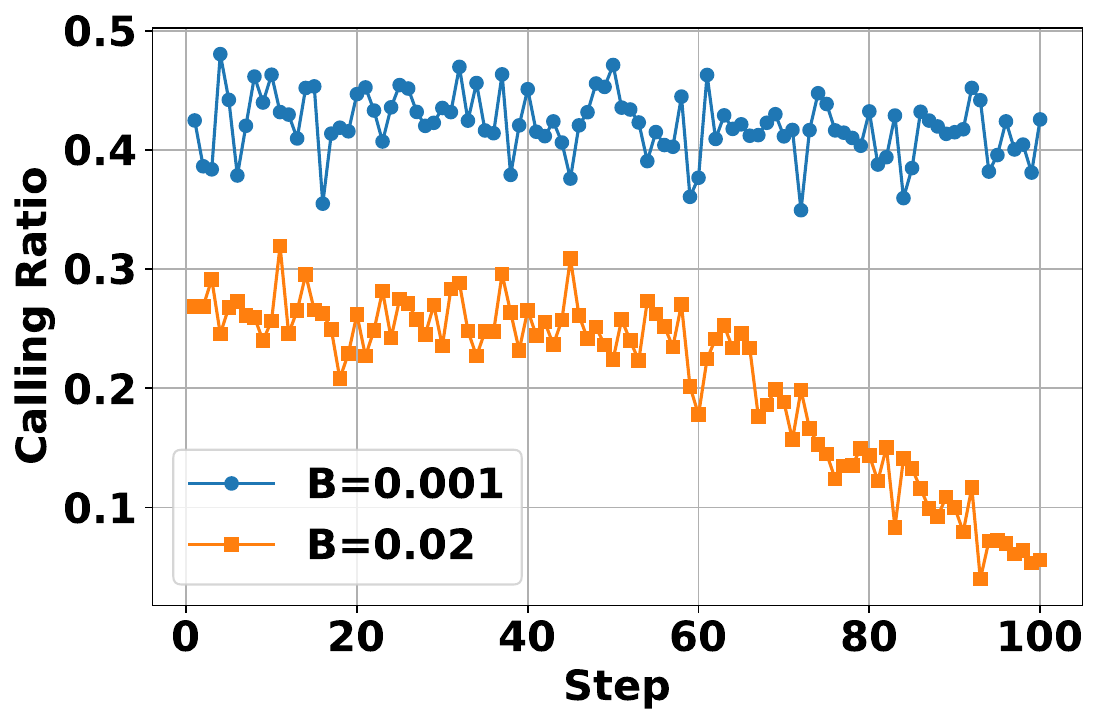}
        \caption{GPT-4.1}
        \label{fig:llm_ratio_gpt41}
    \end{subfigure}\hfill
    \begin{subfigure}[b]{0.32\textwidth}
        \centering
        \includegraphics[width=\textwidth]{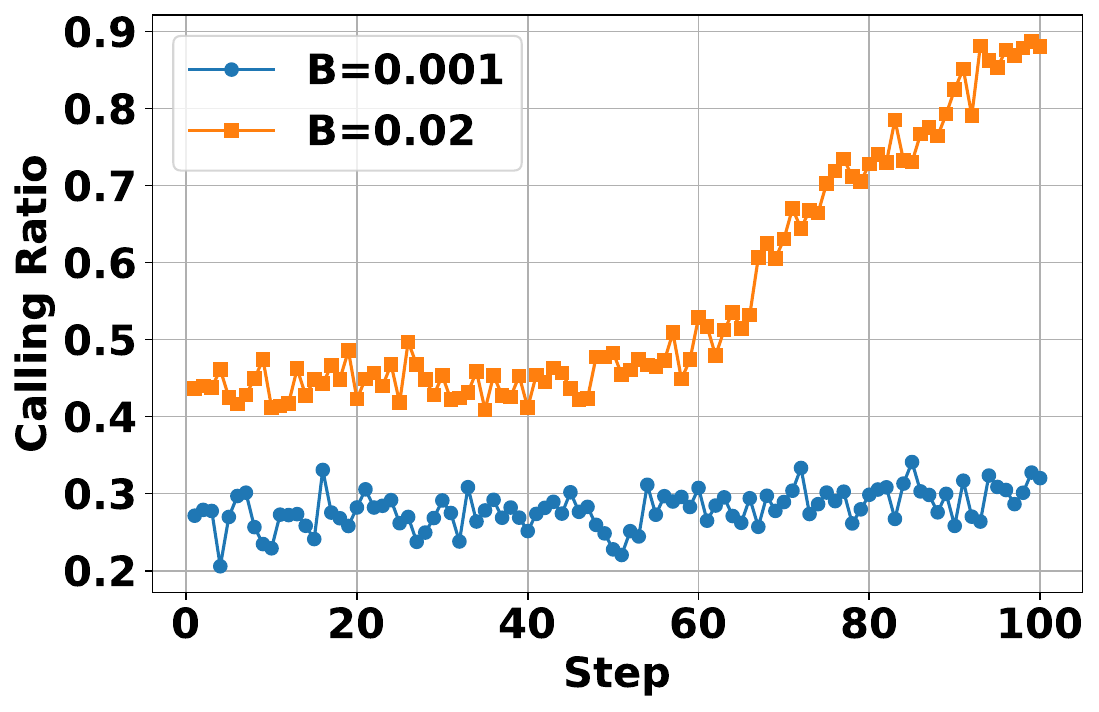}
        \caption{o3}
        \label{fig:llm_ratio_o3}
    \end{subfigure}
    \caption{\textbf{Calling ratio of different expert LLMs during training.} Performance and cost ranking: o3 $>$ GPT-4.1 $>$ GPT-4.1-nano. (1) Under a high budget threshold (\( B = 0.02 \)), the controller increasingly prioritizes o3 as training progresses. (2) Under a low budget threshold (\( B = 0.001 \)), the system avoids over-reliance on o3 despite its stronger performance.}
    \label{fig:llm-ratio-study}
\end{figure*}

\begin{figure*}[t]
    \centering
    \begin{subfigure}[b]{0.32\textwidth}
        \centering
        \includegraphics[width=\textwidth]{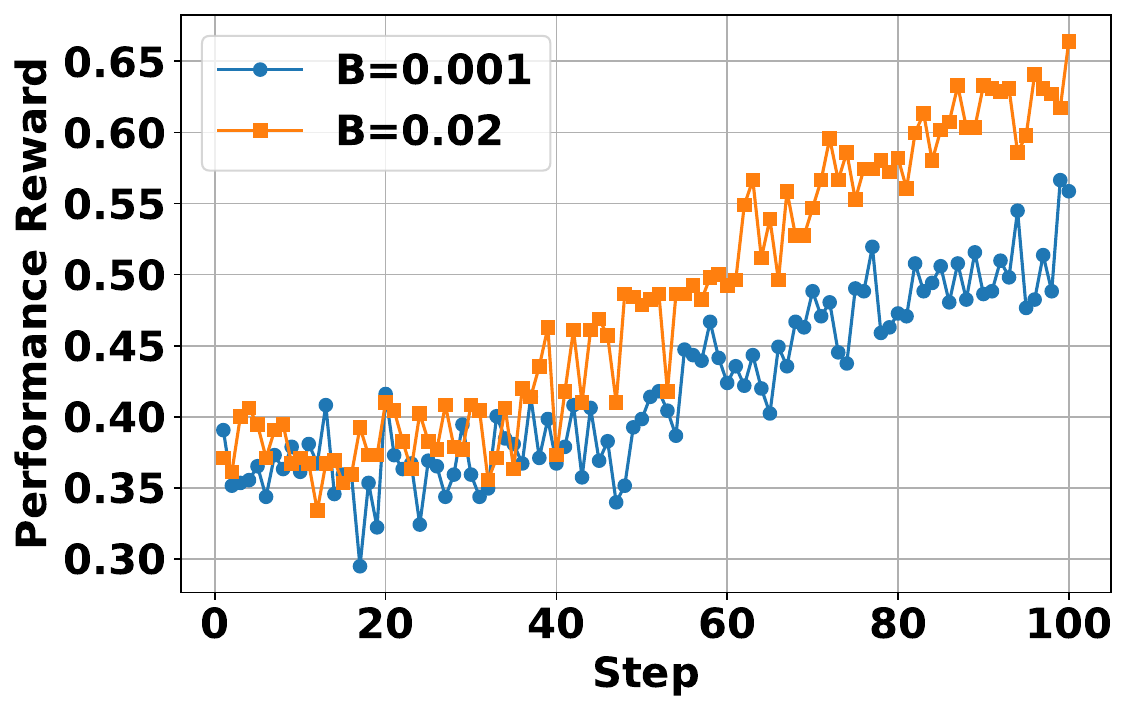}
        \caption{Performance Reward $r_p(\bm{x}, \bm{y})$}
        \label{fig:task_reward}
    \end{subfigure}\hfill
    \begin{subfigure}[b]{0.32\textwidth}
        \centering
        \includegraphics[width=\textwidth]{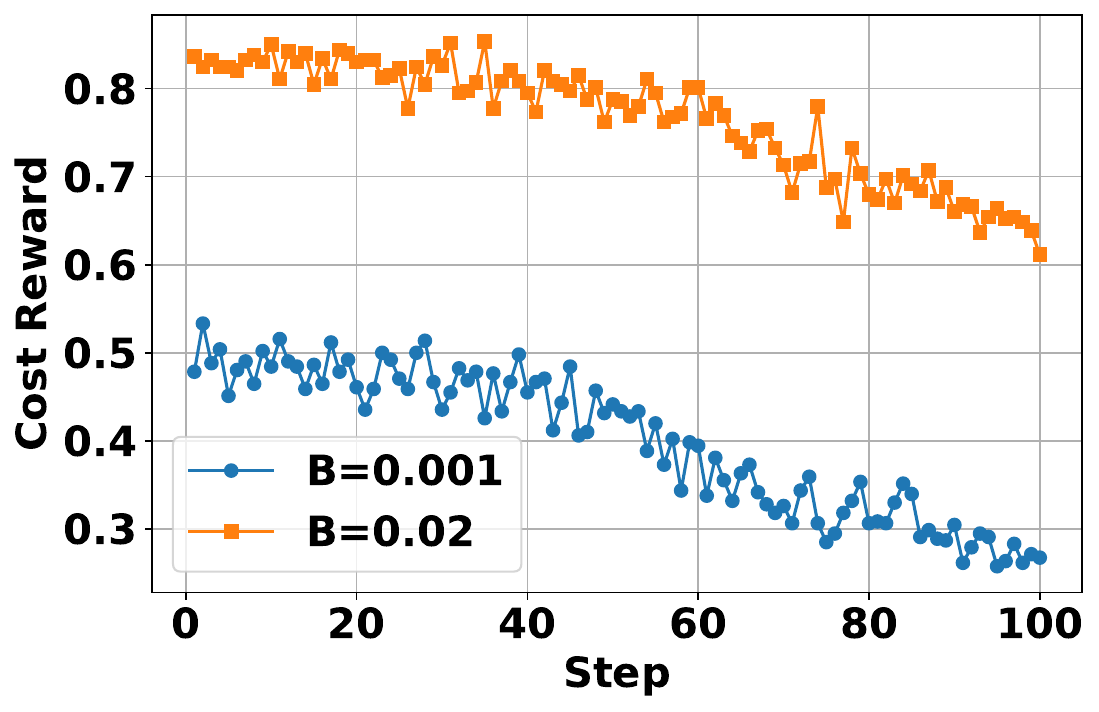}
        \caption{Cost Reward $r_c(\bm{y})$}
        \label{fig:price_reward}
    \end{subfigure}\hfill
    \begin{subfigure}[b]{0.32\textwidth}
        \centering
        \includegraphics[width=\textwidth]{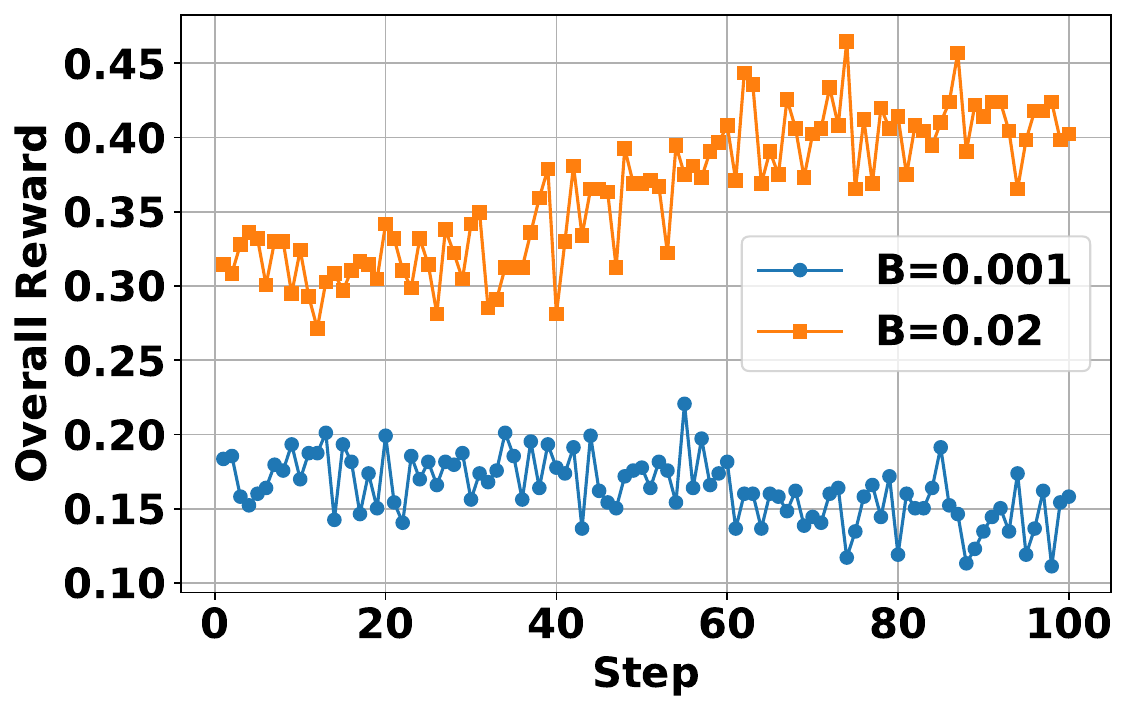}
        \caption{Overall Reward $r_{\phi}(\bm{x}, \bm{y})$}
        \label{fig:overall_reward}
    \end{subfigure}
    \caption{\textbf{Training rewards.} (1) The performance reward \( r_p(\bm{x}, \bm{y}) \) consistently increases under both budget settings as training progresses. (2) The cost reward \( r_c(\bm{y}) \) decreases over time, as the system hits the budget threshold more frequently when performance improves. (3) The overall reward \( r_{\phi}(\bm{x}, \bm{y}) \) rises steadily for \( B = 0.02 \) but fluctuates for \( B = 0.001 \), reflecting the interplay between performance and cost.}
    \label{fig:reward-score}
\end{figure*}

\subsection{Ratio of Expert LLM Calls for Different Budgets Mode}

We further analyze the ratio of expert LLM calls made by the controller under different budget levels. In this experiment, we adopt a two-LLM system, with Qwen2.5-7B-Instruct as the controller and o3 as the expert LLM, and consider three budget levels: low, medium, and high. We examine two styles of system prompts, detailed in Appendix~\ref{apx:two-llm-prompt} and \ref{apx:four-llm-prompt}. The key difference lies in how budget constraints are expressed:
Prompt A explicitly enforces a hard constraint, instructing the controller not to call the expert in the low-budget mode and to always call the expert in the high-budget mode.
Prompt B imposes no hard constraints but encourages the controller to avoid expert calls in the low-budget mode and to prefer expert calls in the high-budget mode.

Figure~\ref{fig:llm-ratio} presents the results. We observe that:
(1) For both prompt types, the expert-call ratio follows the expected order of low < medium < high, indicating that RL successfully learns budget-dependent behaviors;
(2) As training progresses, the overall ratio of expert calls gradually increases, since o3 generally outperforms Qwen2.5-7B, and the reward function incentivizes calling the expert more frequently to maximize performance;
(3) The system prompt design affects training dynamics. With the constrained Prompt A, the controller strictly follows the rules and does not explore expert calls in low-budget mode throughout training. In contrast, with the more flexible Prompt B, the controller learns to occasionally call experts even under low-budget mode, leading to a higher expert utilization ratio.

\section{Understanding the Cost-controlled RL dynamics}

In this section, we verify the effectiveness of our performance–cost reward design through controlled experiments. 
To eliminate the influence of other factors, we fix the budget size $B$ for all training samples and focus on a single budget level. 
We compare the learning dynamics of systems trained with $B=0.001$ and $B=0.02$, respectively. 
The setup involves a trained controller (Qwen2.5-7B-Instruct) and three expert LLMs: GPT-4.1, GPT-4.1-nano, and o3, allowing us to examine how different budget allocations affect the controller’s behavior during training.

\begin{figure}[t]
    \centering
    \includegraphics[width=0.35\textwidth]{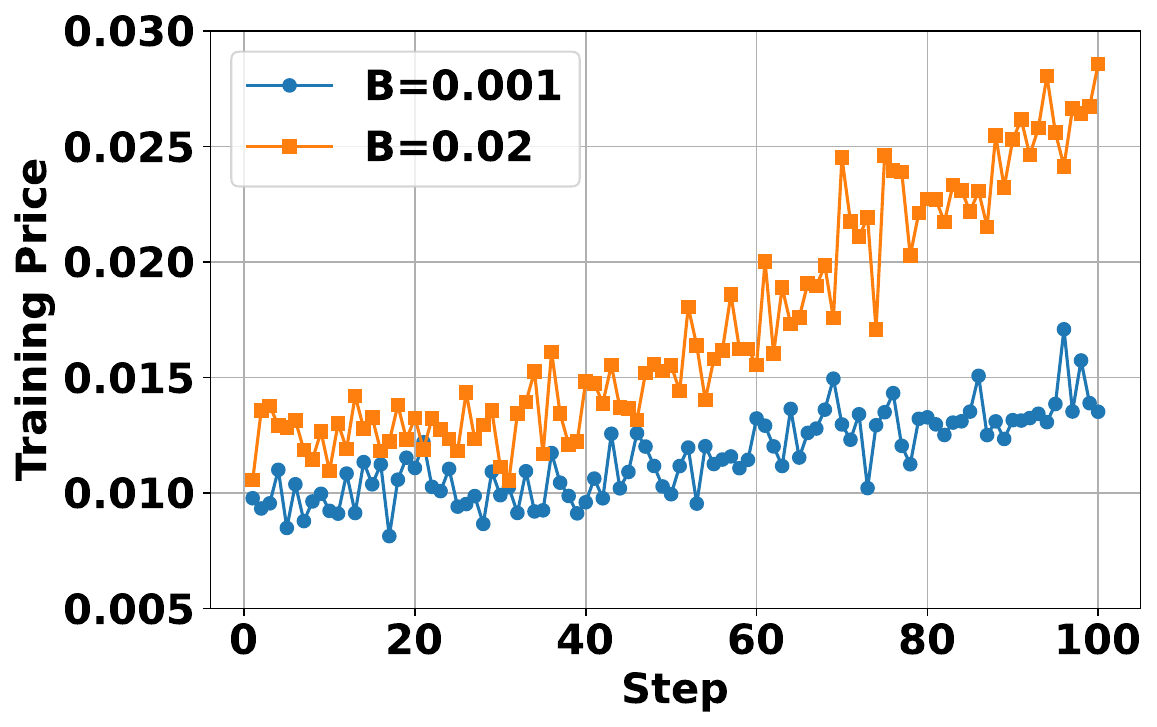} 
    \caption{\textbf{Training price.} (1) The per-query cost increases over the course of training under both budget settings. (2) With a higher budget (\( B = 0.02 \)), the per-query cost of the multi-LLM system is consistently higher than that under a lower budget (\( B = 0.001 \)).}
    \label{fig:train-price} 
\end{figure}

\begin{figure*}[t]
    \centering
    \begin{subfigure}[b]{0.24\textwidth}
        \centering
        \includegraphics[width=\textwidth]{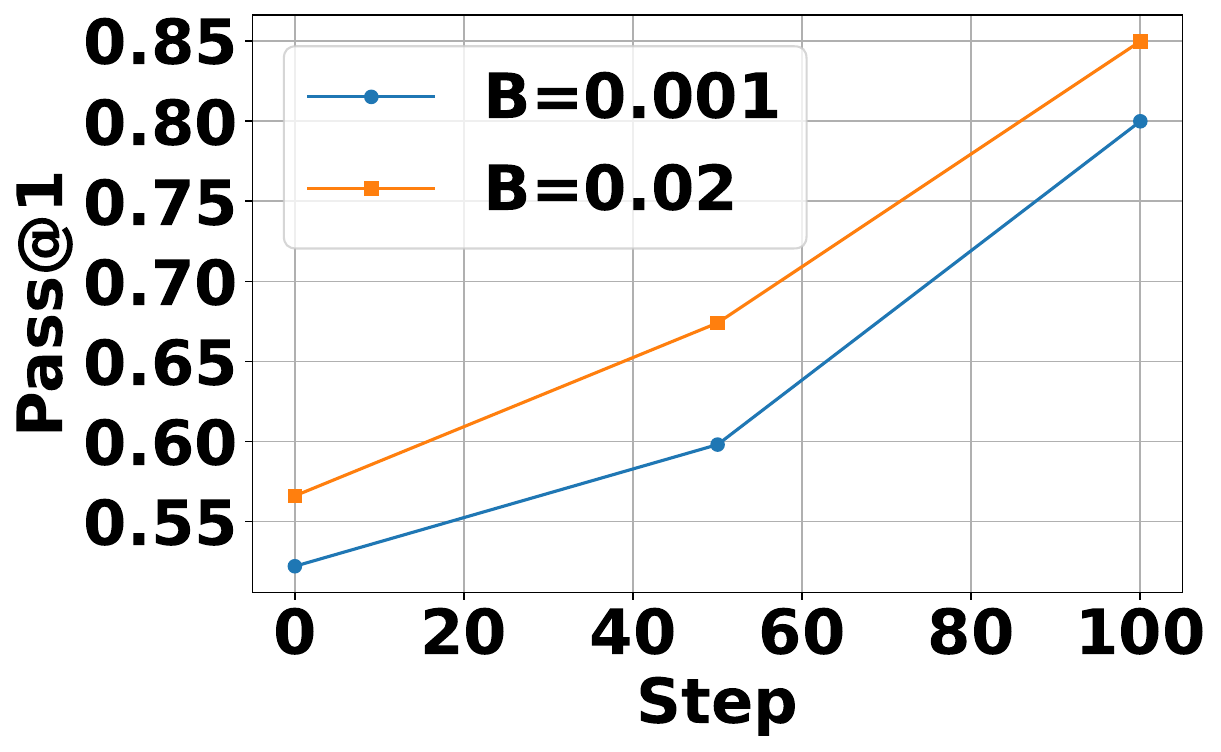}
        \caption{MATH Pass@1}
        \label{fig:math_task}
    \end{subfigure}
    \hfill
    \begin{subfigure}[b]{0.24\textwidth}
        \centering
        \includegraphics[width=\textwidth]{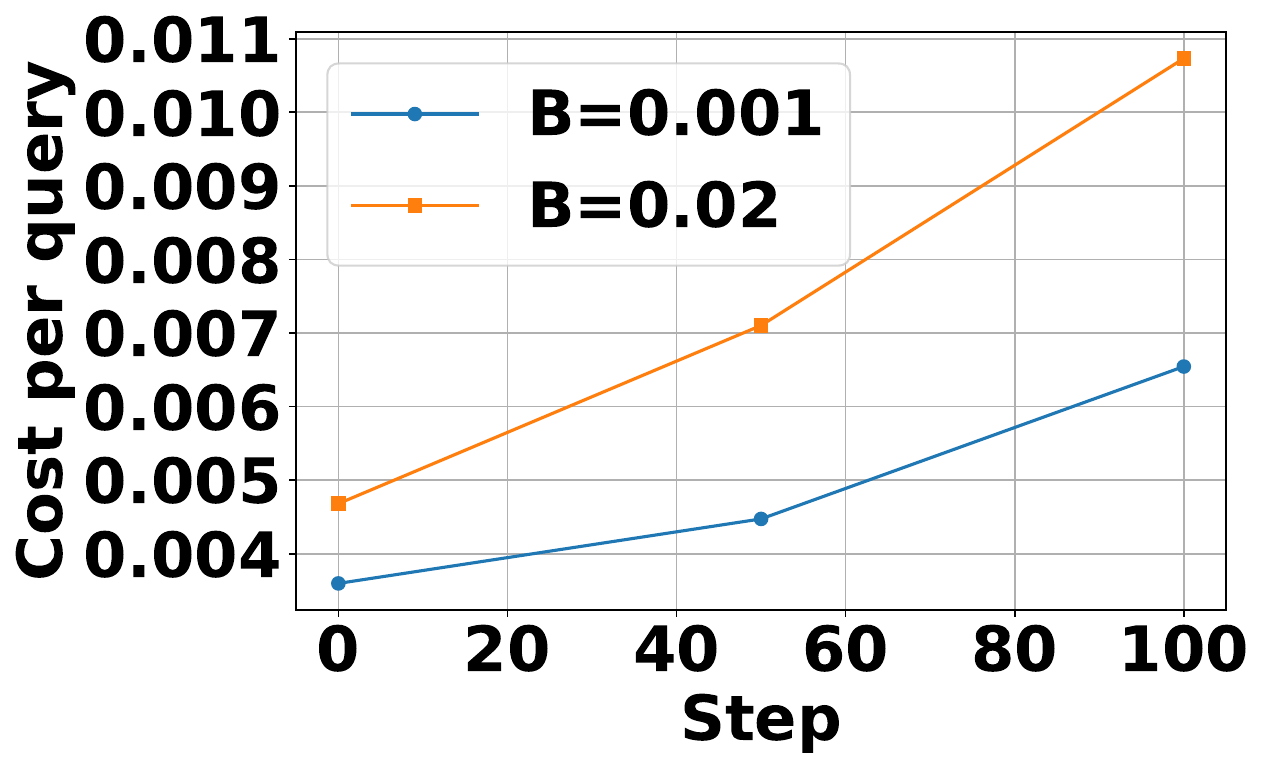}
        \caption{MATH Cost}
        \label{fig:math_price}
    \end{subfigure}
    \hfill
    \begin{subfigure}[b]{0.24\textwidth}
        \centering
        \includegraphics[width=\textwidth]{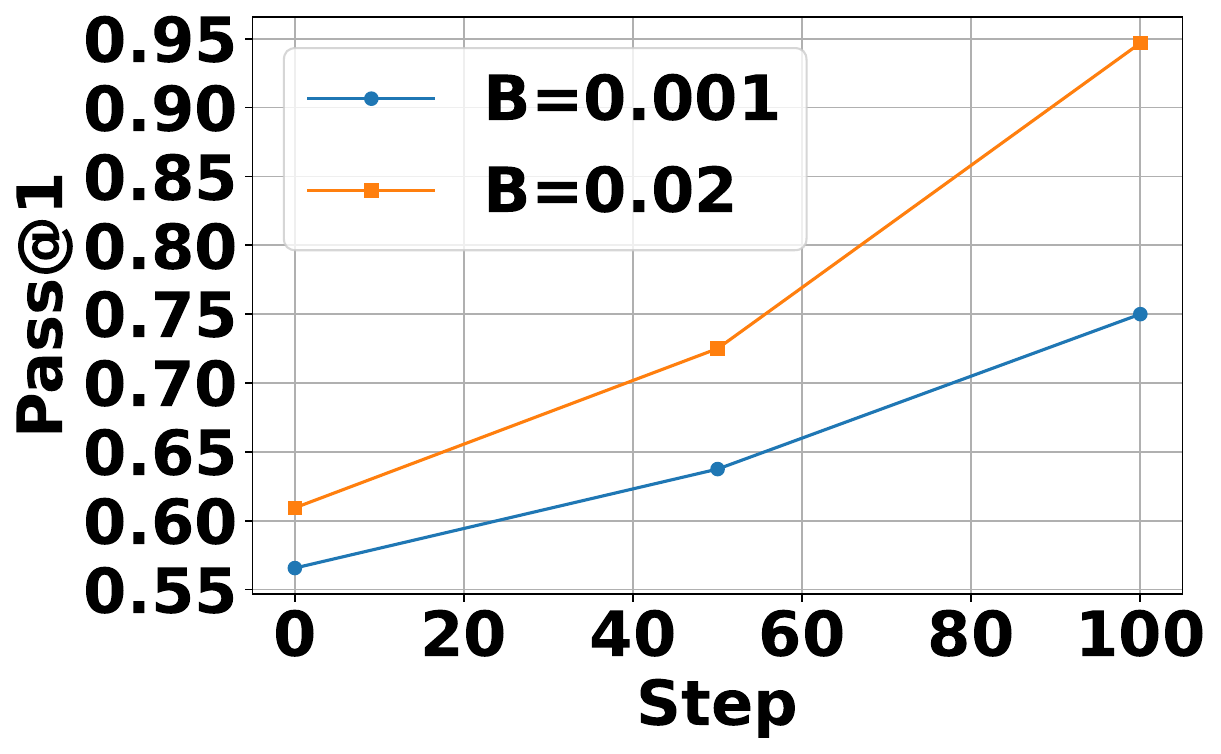}
        \caption{AMC Pass@1}
        \label{fig:amc_task}
    \end{subfigure}
    \hfill
    \begin{subfigure}[b]{0.24\textwidth}
        \centering
        \includegraphics[width=\textwidth]{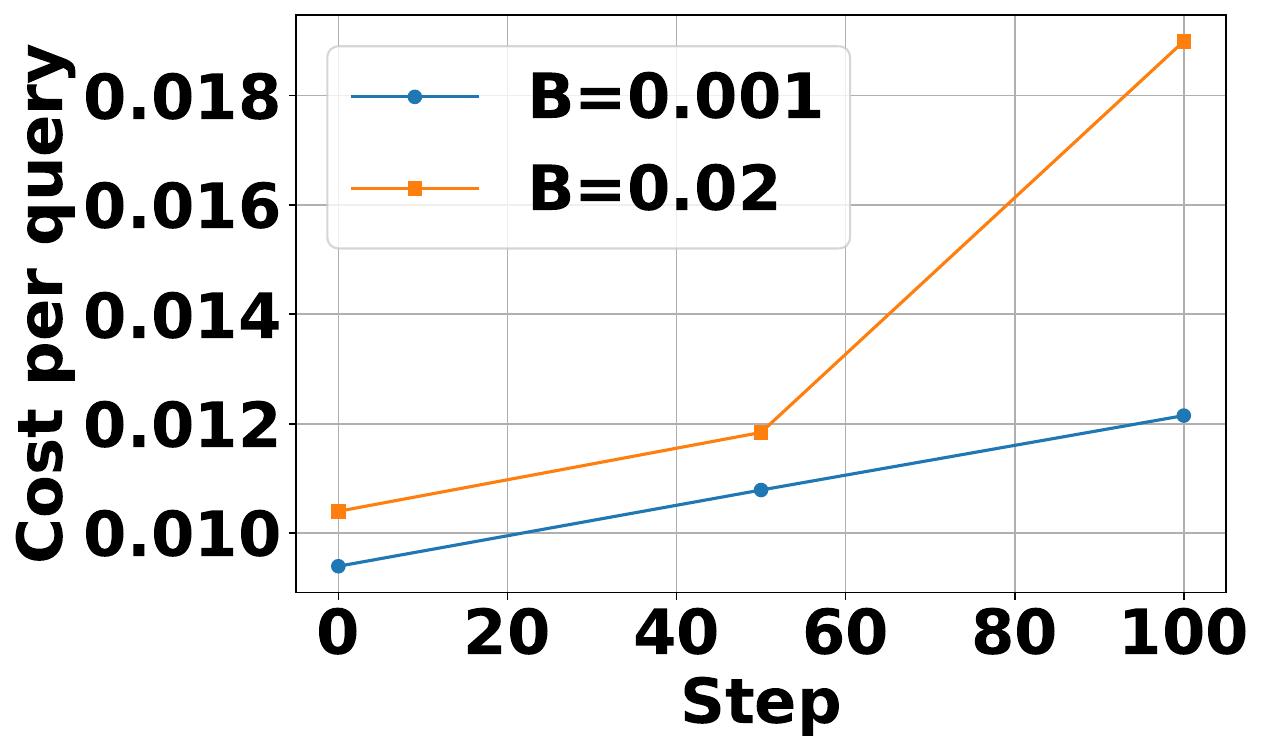}
        \caption{AMC Cost}
        \label{fig:amc_price}
    \end{subfigure}

    \caption{\textbf{Testing behavior.} (1) Systems trained with a higher budget threshold (\( B \)) achieve better task performance than those trained with a lower \( B \). (2) They also incur higher inference costs. This shows that the budget-dependent behaviors learned during training generalize to unseen data.}
    \label{fig:test_behavior}
\end{figure*}

\paragraph{Ratio of Expert LLM Calls.}  
We first examine how the ratio of different expert LLM calls evolves during RL training. Figure~\ref{fig:llm-ratio-study} presents the results. 
(1) Under a \textit{high budget threshold} (\( B = 0.02 \)), the controller increasingly prioritizes o3 over other experts as training progresses. This behavior arises because calls to o3 typically yield higher task rewards \( r_p(\bm{x}, \bm{y}) \). With a sufficiently large \( B \) such that \( c(\bm{y}) \leq B \), prioritizing o3 leads to a higher overall reward \( r_{\phi}(\bm{x}, \bm{y}) \). 
(2) Under a \textit{low budget threshold} (\( B = 0.001 \)), the system does not over-rely on o3. Although o3 can produce higher task rewards, its inference cost often exceeds the budget (\( c(\bm{y}) > B \)), resulting in a final reward of zero. Consequently, the controller learns not to overuse the most powerful but expensive expert. 
These results demonstrate the \textit{effectiveness of our performance--cost reward design} in shaping expert selection behavior under different budget constraints.

\paragraph{Training Reward Dynamics.}  
We further analyze the dynamics of the \textit{training reward}, including the overall reward \( r_{\phi}(\bm{x}, \bm{y}) \), the performance reward \( r_p(\bm{x}, \bm{y}) \), and the cost reward \( r_c(\bm{y}) \), as RL training progresses. Figure~\ref{fig:reward-score} presents the results. 
(1) As training proceeds, the performance reward \( r_p(\bm{x}, \bm{y}) \) consistently increases, regardless of the budget setting \( B \), indicating that the system continuously improves its task performance.
(2) The cost reward \( r_c(\bm{y}) \) decreases over time, showing that as the system becomes stronger, it hits the budget constraint more frequently. Moreover, a smaller budget threshold \( B \) typically leads to a lower cost reward.  
(3) The overall reward \( r_{\phi}(\bm{x}, \bm{y}) \) increases steadily when \( B = 0.02 \) but exhibits larger fluctuations when \( B = 0.001 \). 
This behavior reflects the interplay between performance and cost: under a fixed budget, improving task performance often relies on calling more expensive expert LLMs, which in turn reduces the cost reward.

\paragraph{Training Price.}  
We analyze how the \textit{per-query cost} evolves during RL training. Figure~\ref{fig:train-price} presents the results.  
(1) The per-query cost gradually increases under both budget settings. As training progresses, the controller learns to call expert LLMs more frequently to obtain higher task performance rewards, which naturally leads to increased overall training costs.  
(2) With a higher budget (\( B = 0.02 \)), the per-query cost of the multi-LLM system is consistently higher than that under a lower budget (\( B = 0.001 \)). This demonstrates the effectiveness of our cost-aware reward design in controlling system behavior according to different budget levels.



\paragraph{Testing Behavior.}  
In the previous sections, we have analyzed the system's behavior during training and observed expected learning dynamics. Here, we examine the behavior of the trained multi-LLM system on \textit{unseen testing data}, focusing on two aspects: \textit{performance} and \textit{cost}. Figure~\ref{fig:test_behavior} shows the results on the MATH 500 and AMC 2023 datasets.  
(1) Systems trained with a higher budget threshold (\( B \)) achieve better task-solving performance compared to those trained with a lower \( B \).  
(2) Similarly, systems trained with a higher \( B \) incur higher inference costs than those trained with a smaller \( B \).  
These findings demonstrate that the budget-dependent behaviors learned during training generalize well to unseen data, confirming the effectiveness of our training framework.


\section{Conclusion}  
We presented a centralized reinforcement learning framework for training \textit{cost-efficient and cost-controllable multi-LLM systems}, where a controller LLM selectively coordinates expert models under different budget constraints. By jointly optimizing task performance and query cost, the controller learns budget-aware strategies that adaptively balance when to rely on external experts.
Our analysis highlights three key findings. First, a well-designed performance–cost objective encourages judicious use of expensive experts while maintaining strong overall performance. Second, subtle changes in prompts or budget settings can significantly affect exploration dynamics and routing behavior. Third, the learned strategies generalize to unseen data, achieving favorable performance–cost trade-offs across budget levels.
Future work includes developing more robust exploration strategies, adaptive communication mechanisms, and jointly optimizing all LLMs within the system.





\bibliography{custom}

\appendix

\section{Experimental Settings}\label{apx:setting}

We conduct experiments with Qwen2.5-7B-Instruct \cite{team2024qwen2} as the controller LLM and three GPT family models as the external (expert) LLMs: O3, GPT-4.1, and GPT-4.1-nano.
The strength of these three LLMs in math reasoning tasks is in the order of: O3 > GPT-4.1 > GPT-4.1-nano.
However, the price of solving the problem with the three LLMs is in the same order: O3 > GPT-4.1 > GPT-4.1-nano.

We set the learning rate of the policy LLM to 1e-6 and that of the value LLM to 1e-5. Training is conducted for 500 steps, with warm-up ratios of 0.285 and 0.015 for the policy and value models, respectively. We use Generalized Advantage Estimation (GAE) with parameters $\lambda = 1$ and $\gamma = 1$.

Training is performed on a single node with 8 A100 GPUs. We use a total batch size of 512, with a mini-batch size of 256 and a micro-batch size of 64. The maximum sequence length is set to 4,096 tokens, with a maximum response length of 500 and a maximum length of 500 tokens for retrieved content. To optimize GPU memory usage, we enable gradient checkpointing and use Fully Sharded Data Parallel (FSDP) with CPU offloading.

For efficient LLM rollouts, we adopt vLLM\footnote{\url{https://docs.vllm.ai/en/latest/}} with a tensor parallel size of 1 and GPU memory utilization ratio of 0.6. The rollout sampling uses a temperature of 1.0 and a top-p value of 1.0. The KL divergence regularization coefficient $\beta$ and clip ratio $\epsilon$ are set to 0.001 and 0.2.

We also use gradient checkpointing, FSDP offloading, and vLLM-based rollouts with the same hyperparameters as above. The rollout temperature and top-p values are both set to 1.0, and the KL divergence coefficient $\beta$ and clip ratio $\epsilon$ are fixed at 0.001 and 0.2.

For the single expert LLM experiment, the per-query cost budgets $B$ are set to 0.001 and 1000 for the low and high-budget modes, respectively.
While for the multiple expert LLM study, the per-query cost budgets $B$ are set to 0.001, 0.006, and 1000 for the low-, medium-, and high-budget modes, respectively.

\section{System Prompts}

\subsection{Two LLM System}\label{apx:two-llm-prompt}

\begin{figure}[t]
\centering
\makebox[0.98\columnwidth][c]{%
\begin{tcolorbox}[colback=white,colframe=viridis3!80,title={\bf System Prompt for Low Budget Mode}, width=0.98\columnwidth] %

\footnotesize

You are required to solve the given math problem using a structured, cost-aware multi-LLM controller process.

Inputs: a question, available LLMs, and a price level in [low, medium, high].

Always perform your reasoning within <think> and </think> whenever a new problem is presented or new information is given.

Inside <think>, you must:
1. First, identify the current price level: low, medium, or high
2. Based on the price level, follow the specific routing policy below
3. Do not directly compute the final answer

Price-aware routing policies:

low budget policy:
- You are operating under a low budget constraint
- You must solve all problems yourself, regardless of difficulty
- Do not make any external LLM calls under any circumstances
- Even for complex problems, work within your own capabilities

Self-solve approach (low budget - mandatory):
- Analyze the problem step by step
- Apply mathematical principles and reasoning
- After <think>, provide your solution in <answer> tags only

External LLMs are available but off-limits for low budget: \textcolor{red}{\{available LLMs\}}

Here is some information about the LLMs (for reference only - do not use):
\textcolor{red}{\{expert LLM info\}}

Finalization:
Provide the final solution inside:

<answer> your final answer here </answer>

The answer must be in the simplest, most precise form, with no extra commentary.

Important:
- You are under a low budget - never make external calls
- Solve everything yourself using your own reasoning capabilities
- Focus on accuracy within your computational limits

**Question:** \textcolor{red}{\{query\}}

\end{tcolorbox}
}
\end{figure}

\makebox[0.98\columnwidth][c]{%
\begin{tcolorbox}[colback=white,colframe=viridis3!80,title={\bf System Prompt for Medium Budget Mode}, width=0.98\columnwidth] %

\footnotesize

You are required to solve the given math problem using a structured, cost-aware multi-LLM controller process.

Inputs: a question, available LLMs and a price level in [low, medium, high].

Always perform your reasoning within <think> and </think> whenever a new problem is presented or new information is given.

Inside <think>, you must:
1. First, identify the current price level: low, medium, or high
2. Based on the price level, follow the specific routing policy below
3. Do not directly compute the final answer

Price-aware routing policies:

medium budget policy:
- You are operating under a medium budget constraint
- Carefully assess the difficulty level of the problem
- Make strategic decisions about when to use external LLMs

Difficulty assessment for medium budget:
- easy/moderate problems: Solve yourself to save costs
- difficult/complex problems: Use external LLMs for better accuracy

Self-solve option (easy/moderate problems):
If the problem involves:
- Basic arithmetic, algebra, or geometry
- Standard mathematical procedures you're confident with
- Problems you can solve step-by-step without uncertainty
Then solve it yourself:
- Do not issue any external calls
- After <think>, provide your solution in <answer> tags only

External-solve option (difficult/complex problems):
If the problem involves:
- Advanced mathematical concepts
- Multi-step reasoning with high complexity
- Areas where you're uncertain about the approach
Then use external LLMs:

<call> LLM <t> your decorated query here </t> </call>

Here are the available external LLMs: \textcolor{red}{\{available LLMs\}}. \
For example: \textcolor{red}{\{calling examples\}}

The external LLM's response will be returned inside: <response> ...content... </response>
For example: \textcolor{red}{\{response examples\}}

Here is some information about the LLMs:
\textcolor{red}{\{expert LLM info\}}

Query decoration guidelines for external calls:
- Include the exact problem statement and any constraints
- Request clear step-by-step reasoning
- Set appropriate token limits for cost control

Finalization:
Once you have all the necessary information (either via self-solve or external calls), provide the final solution inside:

<answer> your final answer here </answer>

The answer must be in the simplest, most precise form, with no extra commentary.

Important:
- You are under a medium budget - be strategic about external calls
- Only use external LLMs for genuinely difficult problems
- Always justify your difficulty assessment in <think>

**Question:** \textcolor{red}{\{query\}}

\end{tcolorbox}
}

\makebox[0.98\columnwidth][c]{%
\begin{tcolorbox}[colback=white,colframe=viridis3!80,title={\bf System Prompt for High Budget Mode}, width=0.98\columnwidth] %

\footnotesize

You are required to solve the given math problem using a structured, cost-aware multi-LLM controller process.

Inputs: a question, available LLMs, and a price level in [low, medium, high].

Always perform your reasoning within <think> and </think> whenever a new problem is presented or new information is given.

Inside <think>, you must:
1. First, identify the current price level: low, medium, or high
2. Based on the price level, follow the specific routing policy below
3. Do not directly compute the final answer

Price-aware routing policies:

high budget policy:
- You are operating under a high budget constraint
- You have ample resources to use external LLMs
- Always delegate to external LLMs for maximum accuracy and reliability
- Do not attempt to solve problems yourself

External-solve approach (high budget - mandatory):
You must use external LLMs for all problems, regardless of difficulty level.

<call> LLM <t> your decorated query here </t> </call>

Here are the available external LLMs: \textcolor{red}{\{available LLMs\}}. \
For example: \textcolor{red}{\{calling examples\}}

The external LLM's response will be returned inside: <response> ...content... </response>
For example: \textcolor{red}{\{response examples\}}

Here is some information about the LLMs:
\textcolor{red}{\{expert LLM info\}}

You may call external LLMs multiple times within budget for:
- Initial problem solving
- Verification of results
- Alternative approaches
- Cross-checking answers

Query decoration guidelines for external calls:
- Include the exact problem statement and any constraints
- Request detailed step-by-step reasoning
- Ask for verification of the final answer
- Set generous token limits to ensure comprehensive responses

Finalization:
Once you receive the external LLM response(s), provide the final solution inside:

<answer> your final answer here </answer>

The answer must be in the simplest, most precise form, with no extra commentary.

Important:
- You are under a high budget - always use external LLMs
- Never solve problems yourself when external LLMs are available
- Leverage the superior capabilities of external models for the best results
- You may make multiple calls for verification if needed

**Question:** \textcolor{red}{\{query\}}

\end{tcolorbox}
}

\subsection{Four LLM System}\label{apx:four-llm-prompt}

\begin{figure}[t]
\centering
\makebox[0.98\columnwidth][c]{%
\begin{tcolorbox}[colback=white,colframe=viridis3!80,title={\bf System Prompt for Low Budget Mode}, width=0.98\columnwidth] %

\footnotesize

You are required to solve the given math problem using a structured, cost-aware multi-LLM controller process.

Inputs: a question, available LLMs, and a price level in [low, medium, high].

Always perform your reasoning within <think> and </think> whenever a new problem is presented or new information is given.

Inside <think>, do not directly compute the final answer. Instead, reason about:
- Whether the task is easy (you can confidently solve it yourself) or non-easy (external assistance recommended). Briefly justify.
- If non-easy, which external LLM(s) are most suitable and why, at the price level.
- How to decorate/format the query you will send externally.

Self-solve option (easy only).
If you decide the task is easy, you may solve it yourself. In this case:
- Do not issue any external calls.
- Add your reasoning about the task itself inside <think>.
- After <think>, provide the final solution in <answer> tags only.

External-solve option (non-easy).
If the task is non-easy, you must feed the problem (or a decorated version) to one or more external LLMs using:

<call> LLM <t> your decorated query here </t> </call>

Here are the available external LLMs: \textcolor{red}{\{available LLMs\}}. \
For example: \textcolor{red}{\{calling examples\}}

The external LLM's response will be returned inside: <response> ...content... </response>
For example: \textcolor{red}{\{response examples\}}

Here is some information about the LLMs:
\textcolor{red}{\{expert LLM info\}}

Price-aware routing policy (applies when calling externals):

Query decoration guidelines for external calls:
- Include the exact problem statement and any constraints.
- Impose token limits appropriate to the price level and set deterministic decoding.
- Always minimize redundant calls, reuse prior results, and escalate only when uncertainty or inconsistency warrants it.

Finalization.
Once you have all the necessary information (either via self-solve or external calls), provide the final solution inside:

<answer> your final answer here </answer>

The answer must be in the simplest, most precise form, with no extra commentary.

Important:
- You are under a low budget - Prioritize solving the task by yourself
- Prioritize not calling the external capable model(s).
- Keep prompts compact; request “brief steps + final answer,” and set strict token limits.
- Never directly compute the final answer inside <think> before analyzing the difficulty of the question.

**Question:** \textcolor{red}{\{query\}}

\end{tcolorbox}
}
\end{figure}

\makebox[0.98\columnwidth][c]{%
\begin{tcolorbox}[colback=white,colframe=viridis3!80,title={\bf System Prompt for Medium Budget Mode}, width=0.98\columnwidth] %

\footnotesize

You are required to solve the given math problem using a structured, cost-aware multi-LLM controller process.

Inputs: a question, available LLMs, and a price level in [low, medium, high].

Always perform your reasoning within <think> and </think> whenever a new problem is presented or new information is given.

Inside <think>, do not directly compute the final answer. Instead, reason about:
- Whether the task is easy (you can confidently solve it yourself) or non-easy (external assistance recommended). Briefly justify.
- If non-easy, which external LLM(s) are most suitable and why, at the price level.
- How to decorate/format the query you will send externally.

Self-solve option (easy only).
If you decide the task is easy, you may solve it yourself. In this case:
- Do not issue any external calls.
- Add your reasoning about the task itself inside <think>.
- After <think>, provide the final solution in <answer> tags only.

External-solve option (non-easy).
If the task is non-easy, you must feed the problem (or a decorated version) to one or more external LLMs using:

<call> LLM <t> your decorated query here </t> </call>

Here are the available external LLMs: \textcolor{red}{\{available LLMs\}}. \
For example: \textcolor{red}{\{calling examples\}}

The external LLM's response will be returned inside: <response> ...content... </response>
For example: \textcolor{red}{\{response examples\}}

Here is some information about the LLMs:
\textcolor{red}{\{expert LLM info\}}

Price-aware routing policy (applies when calling externals):

Query decoration guidelines for external calls:
- Include the exact problem statement and any constraints.
- Impose token limits appropriate to the price level and set deterministic decoding.
- Always minimize redundant calls, reuse prior results, and escalate only when uncertainty or inconsistency warrants it.

Finalization.
Once you have all the necessary information (either via self-solve or external calls), provide the final solution inside:

<answer> your final answer here </answer>

The answer must be in the simplest, most precise form, with no extra commentary.

Important:

- You are under a medium budget - be strategic about external calls
- Use minimal but sufficient context; cap tokens conservatively
- Only use external LLMs for genuinely difficult problems
- Never directly compute the final answer inside <think> before analyzing the difficulty of the question.

**Question:** `{question}`

\end{tcolorbox}
}

\makebox[0.98\columnwidth][c]{%
\begin{tcolorbox}[colback=white,colframe=viridis3!80,title={\bf System Prompt for High Budget Mode}, width=0.98\columnwidth] %

\footnotesize

You are required to solve the given math problem using a structured, cost-aware multi-LLM controller process.

Inputs: a question, available LLMs, and a price level in [low, medium, high].

Always perform your reasoning within <think> and </think> whenever a new problem is presented or new information is given.

Inside <think>, do not directly compute the final answer. Instead, reason about:
- Whether the task is easy (you can confidently solve it yourself) or non-easy (external assistance recommended). Briefly justify.
- If non-easy, which external LLM(s) are most suitable and why, at the price level.
- How to decorate/format the query you will send externally.

Self-solve option (easy only).
If you decide the task is easy, you may solve it yourself. In this case:
- Do not issue any external calls.
- Add your reasoning about the task itself inside <think>.
- After <think>, provide the final solution in <answer> tags only.

External-solve option (non-easy).
If the task is non-easy, you must feed the problem (or a decorated version) to one or more external LLMs using:

<call> LLM <t> your decorated query here </t> </call>

Here are the available external LLMs: \textcolor{red}{\{available LLMs\}}. \
For example: \textcolor{red}{\{calling examples\}}

The external LLM's response will be returned inside: <response> ...content... </response>
For example: \textcolor{red}{\{response examples\}}

Here is some information about the LLMs:
\textcolor{red}{\{expert LLM info\}}

Price-aware routing policy (applies when calling externals):

Query decoration guidelines for external calls:
- Include the exact problem statement and any constraints.
- Impose token limits appropriate to the price level and set deterministic decoding.
- Always minimize redundant calls, reuse prior results, and escalate only when uncertainty or inconsistency warrants it.

Finalization.
Once you have all the necessary information (either via self-solve or external calls), provide the final solution inside:

<answer> your final answer here </answer>

The answer must be in the simplest, most precise form, with no extra commentary.

Important:
- You are under high budget - prioritize external LLMs
- Leverage the superior capabilities of external models for the best results
- Never directly compute the final answer inside <think> before analyzing the difficulty of the question.

**Question:** `{question}`

\end{tcolorbox}
}

\end{document}